\PassOptionsToPackage{prologue,dvipsnames}{xcolor}
\documentclass[fleqn,10pt]{wlscirep}
\usepackage[utf8]{inputenc}
\usepackage[T1]{fontenc}
\usepackage{microtype}
\usepackage{graphicx}
\usepackage{subfigure}
\usepackage{booktabs} 
\usepackage{multirow}
\usepackage[symbol]{footmisc}
\usepackage{colortbl}
\usepackage{amsmath}
\usepackage{amssymb}
\usepackage{mathtools}
\usepackage{amsthm}

\usepackage[capitalize,noabbrev]{cleveref}

\definecolor{lightblue}{rgb}{0.80,0.80,0.97}
\definecolor{lightred}{rgb}{0.97,0.80,0.80}

\title{Rethinking VLMs and LLMs for Image Classification}

\author[1+]{Avi Cooper}
\author[2+]{Keizo Kato}
\author[2,3]{Chia-Hsien Shih}
\author[2]{Hiroaki Yamane}
\author[1]{Kasper Vinken}
\author[2]{Kentaro Takemoto}
\author[2]{Taro Sunagawa}
\author[2]{Hao-Wei Yeh}
\author[1]{Jin Yamanaka}
\author[1*]{Ian Mason}
\author[1*]{Xavier Boix}

\affil[1]{Fujitsu Research of America, Santa Clara, CA, USA}
\affil[2]{Fujitsu Limited, Kawasaki, Japan}
\affil[3]{University of Illinois at Urbana-Champaign
Urbana, IL, USA}

\affil[+]{these authors contributed equally to this work}
\affil[*]{these authors contributed equally to this work}

\begin{abstract}
Visual Language Models~(VLMs) are now increasingly being merged with Large Language Models~(LLMs) to enable new capabilities, particularly in terms of improved interactivity and open-ended responsiveness. While these are remarkable capabilities, the contribution of LLMs to enhancing the longstanding key problem of classifying an image among a set of choices remains unclear. Through extensive experiments involving seven models, ten visual understanding datasets, and multiple prompt variations per dataset, we find that, for object and scene recognition, VLMs that do not leverage LLMs can achieve better performance than VLMs that do. Yet at the same time, leveraging LLMs can improve performance on tasks requiring reasoning and outside knowledge. In response to these challenges, we propose a pragmatic solution: a lightweight fix involving a relatively small LLM that efficiently routes visual tasks to the most suitable model for the task. The LLM router undergoes training using a dataset constructed from more than 2.5 million examples of pairs of visual task and model accuracy. Our results reveal that this lightweight fix surpasses or matches the accuracy of state-of-the-art alternatives, including GPT-4V and HuggingGPT, while improving cost-effectiveness.
\end{abstract}
\begin{document}

\flushbottom
\maketitle
%
%
\thispagestyle{empty}


\section{Introduction} 
\label{sec:intro}

\begin{figure}[tb]
    \centering
    \includegraphics[width=0.5\linewidth]{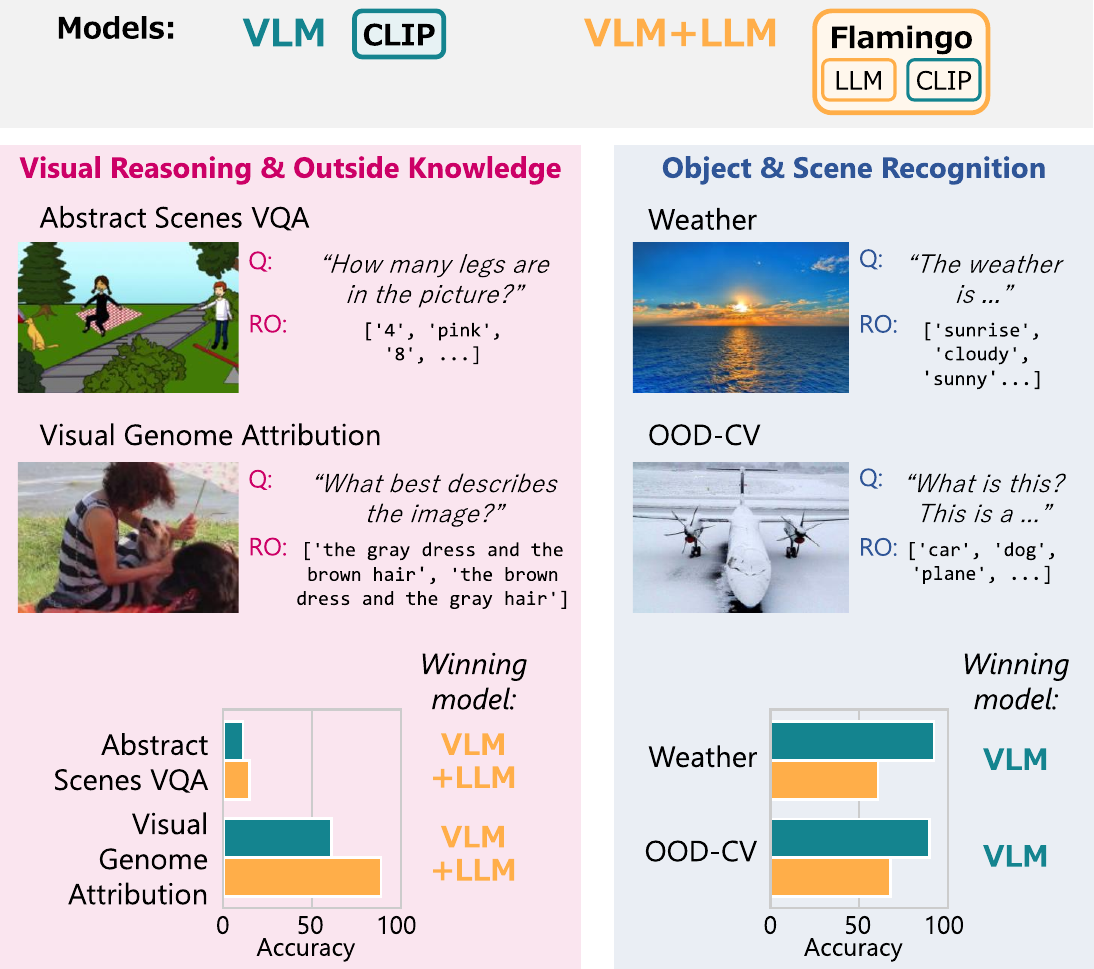}
    \caption{VLMs achieve higher performance on object and scene recognition tasks than VLM+LLMs with the same vision encoder. Q: prompt provided to the models. RO: response options provided to the models. Left: on tasks involving visual reasoning and outside knowledge the VLM+LLM (Flamingo) outperforms the VLM (CLIP) with the same vision encoder. Right: on object and scene recognition the VLM is superior to the VLM+LLM. Note that Flamingo only uses CLIP's vision encoder, but since this vision encoder is pre-trained with the full CLIP architecture, in the figure we show CLIP as part of Flamingo.}
    \label{fig:vlm_vlmplsllm_diff_on_datasets}
\end{figure}

Many of the best methods for visual representation learning now make use of Visual Language Models (VLMs) \cite{radford2021clip, li2022blip, zhai2022lit}. These models combine vision and language (e.g.\ with contrastive training) in order to learn general vision-language representations. Recently, several new VLMs have been developed that combine VLMs with pre-trained Large Language Models (VLM+LLMs)\footnote[2]{We use the term LLM as it is commonly used, i.e.\ an auto-regressive text-generating model trained exclusively on text data. To distinguish VLMs that do and do not leverage an LLM as part of their architecture, we use the terms VLM+LLM and VLM respectively.} \cite{tiong2022pnp, alayrac2022flamingo, chen2023pali}. 
Since LLMs are trained on large corpora of text and computer code, combining pre-trained LLMs with VLMs has brought many benefits to VLMs, including the addition of outside knowledge and reasoning, iterative interaction, and an ability to provide open-ended responses.

While these new abilities are impressive, image classification tasks in which the image is categorized into a closed set of possible choices, such as in classic object recognition, constitute one of the primary goals of VLMs and computer vision in general. 
At the moment, it is unclear whether the additional capabilities of VLM+LLMs also contribute to an overall improvement to image classification.  
In the literature on VLM+LLMs, the evaluation focus has primarily centered on assessing the new abilities introduced by LLMs, i.e.\ open-ended answering and interactivity, with limited attention given to evaluating their impact on classifying the image into a given set of possible choices as it was traditionally done in image classification. Thus, in this work we re-evaluate VLM+LLMs on a diverse set of visual tasks with the goal of answering: when does leveraging LLMs improve model performance in traditional image classification setup?

We evaluate the performance of existing open source VLMs and VLM+LLMs across a range of benchmarks designed to test object and scene recognition as well as visual reasoning and outside knowledge. 
There are three possible outcomes to this experiment that \emph{a priori} are not possible to discern:
\begin{enumerate}
\item VLM+LLMs are superior in all the tasks for image classification. This result may be the most expected, as VLM+LLMs currently dominate state-of-the-art accuracy in most vision tasks. 
\item  VLMs outperform VLM+LLMs depending on datasets. If we observe this, it is worth investigating when VLMs are better than VLM+LLMs.
\item VLMs always surpass VLM+LLMs on image classification tasks. This result sounds contrary to expectations at first. However, we argue this is possible since unnecessary linguistic knowledge may interfere with the visual representations. 
\end{enumerate}
Results show that despite increasing parameter counts, VLM+LLMs are less performant than VLMs on object and scene recognition, while being superior on reasoning and outside knowledge tasks. In a series of experiments, we show that these results are consistent over seven models, ten datasets and multiple prompt variations.   Our results are illustrated with examples in Figure~\ref{fig:vlm_vlmplsllm_diff_on_datasets}.

These observations show that different models excel at different vision tasks.
We therefore introduce a computationally efficient system to route the task to the best model. To do so, we fine-tune GPT-2 to act as a router using the results from our experiments (about 2.5 million examples of pairs of visual task and model accuracy) and evaluate the router on held-out tasks. Because of this specialized training, our approach demonstrates a remarkable effectiveness in held-out visual tasks --- it outperforms HuggingGPT~\cite{shen2023hugginggpt}, and matches the accuracy of GPT-4V~\cite{openai2023gpt4} while improving in cost-effectiveness. Therefore, a lightweight LLM can effectively be trained to arbitrate between models and, based on user input, select one that is most suitable for the task at hand.

\section{Do VLM+LLMs Always Beat VLMs?}
\label{sec:observations}

In this section we analyze the capabilities of VLM+LLMs across multiple image classification tasks. We first introduce the VLMs and VLM+LLMs that we analyze and then introduce the evaluation methodology, which includes an overview of the datasets and prompting strategies used. Finally, we present the findings derived from our analysis.

\subsection{Comparing VLM+LLMs with VLMs}

To fairly compare the effect of leveraging LLMs for vision-language processing, we compare VLM+LLMs and VLMs that have exactly the same visual representations. That is, that use identical vision encoders. In this way we assess the effectiveness of different strategies for handling text, one that uses LLMs and the other that does not.

Note that including an LLM in a VLM necessitates the inclusion of additional machinery around the LLM to integrate it with the vision encoder (fusion modules, training objective, additional data, etc.).
Two prevalent strategies exist in designing VLM+LLM models, namely that visual information can be incorporated into the LLM via embeddings (e.g.\ Flamingo~\cite{alayrac2022flamingo}) or via text descriptions of the image (e.g.\ PNP-VQA~\cite{tiong2022pnp}). We investigate models from each approach.

\noindent \textbf{Flamingo vs. CLIP}
The architecture of Flamingo~\cite{alayrac2022flamingo} consists of the pre-trained and frozen vision encoder from CLIP and an LLM, connected with trainable attention layers. 
Flamingo is trained on a mixture of large-scale multimodal data from the web. We use the open-source OpenFlamingo~\cite{awadalla2023openflamingo} implementation, utilizing CLIP ViT-L/14~\cite{radford2021clip} as the vision encoder and MPT~\cite{MosaicML2023Introducing} as the LLM.
Thus, we directly compare Flamingo with CLIP as they have the same vision encoder but different techniques for handling text. To understand the impact of LLM size on performance, we conducted experiments with Flamingo employing different parameter LLM configurations, including 3 billion (3b) and 9 billion (9b) parameters.

\noindent  \textbf{PNP-VQA vs. BLIP.}
PNP-VQA~\cite{tiong2022pnp} facilitates communication between pre-trained and frozen models through language, without any additional training. First, PNP-VQA uses a VLM, namely BLIP~\cite{li2022blip}, to identify the relevant parts in the image for the task at hand and then generates relevant captions (again, with BLIP) for each of the identified parts. Then an LLM, based on UnifiedQA~\cite{khashabi2020unifiedqa}, provides a final answer by taking into account the prompt along with the captions. Since PNP-VQA makes use of BLIP internally, directly comparing them provides insights into handling text with and without leveraging LLMs.

\noindent  \textbf{LiT vs. CLIP and Flamingo.}
Locked-image text Tuning (LiT)~\cite{zhai2022lit} is a VLM with the same vison encoder as CLIP and Flamingo but a larger text encoder than CLIP. It is a VLM and not a VLM+LLM as the text encoder is trained in a contrastive manner between text and image features, rather than leveraging pre-trained LLMs. We test two variations of LiT text encoder: BERT-base with 110M parameters (LiT-B/16) and BERT-large with 340M parameters~\cite{devlin2018bert} (LiT-L/16),  which are much larger than the text encoder in CLIP (vanilla Transformer with 63M parameters~\cite{radford2021clip}). We include LiT in our evaluation to assess whether the differences in performance between VLMs and VLM+LLMs come from the pre-trained LLM rather than from simply having larger text encoders to handle text.

\subsection{Returning to Closed-Ended Evaluation}
\label{closed-ended}

In language modelling tasks, such as translation and image captioning, evaluation is typically open-ended where models engage in open-ended text generation. On the other hand, image classification is inherently a closed-ended task, requiring a selection from a set of options rather than general description. In the open-ended case, evaluation is unclear because the degree of detail required is somewhat ambiguous. It is not sufficient to say that there is a tree in the image --- the model must decide whether it is an elm, oak or willow tree if the task requires so. Thus, we adopt closed-ended evaluation for all experiments.

The methods we employ to compute closed-ended predictions differ across model types.
For VLMs, the prediction of the model is the closed-ended option whose representation is best aligned with the image representation through a dot product. For VLM+LLMs, like PNP-VQA and Flamingo, the logits of each of the closed-ended options are used to compute a sequence probability. The sequence of logits with the highest average probability is the prediction of the model. As we show below in the results, differences in how models handle closed-ended prediction does not affect the overall conclusions.

\begin{table*}[ht]
\begin{scriptsize}
\centering
\setlength{\tabcolsep}{5pt}
\begin{tabular}{ccccccccccc}
\toprule
\textbf{} & \textbf{CIFAR-100} & \textbf{OOD-CV} & \textbf{Weather} & \textbf{\begin{tabular}[c]{@{}c@{}}Skin\\Cancer\end{tabular}} & \textbf{\begin{tabular}[c]{@{}c@{}}Hateful\\Memes\end{tabular}} & \textbf{ScienceQA} & \textbf{\begin{tabular}[c]{@{}c@{}}Visual\\Genome\\Attribution\end{tabular}} & \textbf{\begin{tabular}[c]{@{}c@{}}Visual\\Genome\\Relation\end{tabular}} & \textbf{\begin{tabular}[c]{@{}c@{}}Abstract\\Scenes\\VQA\end{tabular}} & \textbf{\begin{tabular}[c]{@{}c@{}}Binary\\Abstract\\Scenes\end{tabular}} \\
\cmidrule(r){1-5}\cmidrule(l){6-11}
\textbf{Chance Rate} & 1.0 & 16.1 & 31.9 & 57.1 & \textbf{58.7} & 36.0 & 50.1 & 50.2 & 5.72 & 50.8 \\
\cmidrule(r){1-5}\cmidrule(l){6-11}
\rowcolor{lightblue}
\textbf{\scalebox{.7}{VLM} BLIP} & \underline{64.4} / \underline{66.6} & \underline{84.7} / \underline{89.2} &  \underline{80.2} / \underline{86.7} & \underline{56.1} / 57.1 & \underline{48.4} / 53.0 & 39.3 / 40.2 & \underline{79.4} / \underline{79.6} & 54.9 / 56.8 & 16.6 / 19.2 & 49.3 / 49.6 \\
\rowcolor{lightblue}
\textbf{\scalebox{.7}{VLM+LLM} PNP-VQA} & 51.8 / 54.7 & 75.3 / 78.1 & 58.7 / 73.5 & 50.1 / 57.1 & 46.6 / \underline{54.3} &\underline{49.8} / \underline{50.9} & 77.9 / 77.9 & \underline{58.2} / \underline{58.2} & \underline{\textbf{42.0}} / \underline{\textbf{50.5}} & \underline{\textbf{57.8}} / \underline{\textbf{58.1}} \\
\rowcolor{lightred}
\textbf{\scalebox{.7}{VLM} CLIP} & \underline{60.1} / \underline{70.4} & \underline{\textbf{89.4}} / \underline{\textbf{90.9}} & \underline{\textbf{92.0}} / \underline{\textbf{95.1}} & 54.7 / 57.1 & \underline{51.4} / 57.9 & 47.2 / 49.3 & 61.0 / 62.6 & 54.1 / 54.4 & 11.1 / 13.6 & \underline{51.4} / \underline{51.9} \\
\rowcolor{lightred}
\textbf{\scalebox{.7}{VLM+LLM} Flamingo-3B} & 48.3 / 68.1 & 66.0 / 88.4 & 54.4 / 82.7 & 57.2 / 57.1 & 50.0 / \underline{\textbf{58.7}} & \underline{\textbf{50.6}} / \underline{\textbf{52.8}} & 87.0 / 88.3 & 60.3 / 60.6 & \underline{22.2} / \underline{31.5} & 49.2 / 49.2 \\
\rowcolor{lightred}
\textbf{\scalebox{.7}{VLM+LLM} Flamingo-9B} & 53.2 / 64.3 & 72.8 / 87.7 & 64.2 / 87.2 & \underline{\textbf{58.2}} / \underline{\textbf{62.2}} & 50.0 / \underline{\textbf{58.7}} & 49.2 / 51.3 & \underline{\textbf{89.0}} / \underline{\textbf{90.8}} & \underline{\textbf{60.6}} / \underline{\textbf{61.2}} & 14.6 / 18.9 & 49.2 / 49.2 \\
\rowcolor{lightgray}
\textbf{\scalebox{.7}{VLM} LiT-B/16} &\textbf{74.8} / 77.8 & 86.1 / 87.6 & 87.8 / 91.6 & 54.8 / 57.1 & 50.7 / 53.6 & 37.2 / 41.2 & 65.5 / {66.5} & 51.9 / 52.1 & 6.9 / 8.0 & 49.5 / 50.2 \\
\rowcolor{lightgray}
\textbf{\scalebox{.7}{VLM} LiT-L/16} & 74.3 / \textbf{78.4} & 83.9 / 85.3 & 88.3 / 92.0 & {55.3} / \textbf{62.2} & 52.5 / {56.3} & 37.4 / 40.1 & 65.1 / 65.2 & 53.2 / 53.2 & 8.2 / 9.7 & 49.6 / 49.9 \\
\bottomrule
\end{tabular}
\caption{Accuracies of VLMs and VLM+LLMs across datasets (left: average across prompts / right: best prompt accuracy). A VLM+LLM and its constituent VLM are color-coded, with the result for the better performing model underlined for each dataset. The highest accuracy achieved for each dataset is in bold. The first four datasets are object and scene recognition where VLM models are almost always superior to VLM+LLMs. The final six columns show tasks requiring reasoning and outside knowledge on which VLM+LLMs outperform VLMs.}
\label{tab:classification_accuracy}
\end{scriptsize}
\end{table*}

\subsection{Zero-Shot Visual Tasks}

LLMs excel at zero-shot tasks~\cite{brown2020gpt3} and can capture nuanced relationships and contextual information that might not be explicitly present in visual data. Conceivably, the extensive knowledge encapsulated by LLMs could be harnessed to address visual tasks when there is a lack of training data. 

Zero-shot visual tasks allow us to assess the effectiveness of the synergy between VLMs and LLMs in a more controlled manner than if there would be task-specific training samples for fine-tuning. To avoid adding confounding factors arising from fine-tuning design choices (like deciding which layers to unfreeze) all experiments in this paper revolve around zero-shot visual tasks. We next introduce the 10 zero-shot datasets we leverage for our analysis, each characterized by varying degrees of reasoning complexity and knowledge requirements (see Appendix~\ref{app:datasets} for further details).

\emph{CIFAR-100}~\cite{krizhevsky2009learning}--a widely used benchmark for general object recognition. \emph{OOD-CV}~\cite{oodcv}--an out-of-distribution object classification, where objects undergo diverse pose, shape, occlusion and texture variations. \emph{Weather}~\cite{weather}--a multi-class weather recognition from  images. \emph{Skin Cancer}~\cite{skin_cancer}--diagnosis for melanoma detection. \emph{Hateful Memes}~\cite{HatefulMemes}--hatred identification in memes through a combination of image and text analysis. \emph{ScienceQA}~\cite{ScienceQA}--contains questions with images about scientific topics that require outside knowledge and reasoning capabilities. It resembles a high-school science exam. \emph{Visual Genome Relation}~\cite{VGAR}--given an image and a relation of the form ``X relation Y'', this benchmarks
test whether the model can pick the correct order ``X relation Y'', instead of ``Y relation X''. \emph{Visual Genome Attribution}~\cite{VGAR}--evaluates the ability to attribute properties to objects appropriately, structured in a similar fashion as the Visual Genome Relation dataset. \emph{Abstract Scenes VQA}~\cite{Antol_2015_ICCV}--a commonly used dataset to assess the ability to answer questions related to high-level semantic and abstract information from scenes. \emph{Binary Abstract Scenes}~\cite{balanced_binary_vqa}--the same as Abstract Scenes VQA but the response options are restricted to `yes' and `no' in order to remove biased responses.

Note that we do not use  ImageNet or Visual Genome for object recognition datasets as some VLMs use them for training their vision encoders and it is unclear whether these datasets can be effectively used to assess zero-shot capabilities.

\subsection{Prompt Choice}
\label{prompting_choice}
It is well known that prompt choice can have a significant impact on model performance \cite{reynolds2021prompt, zhou2023promptengineering, yang2022empirical, kojima2022reasoners, wang2022self-consistency, pitis2023boosted}, and work has been conducted to specifically improve LLM's reasoning capabilities, through Chain-of-Thought \cite{CoT}, Tree of Thought \cite{tot} and similar processes.

During evaluation we observed that varying the prompt dramatically affected the evaluation performance of the various networks.
For example, when evaluating CLIP on CIFAR-100, using prompts of the form ``What is this? This is aquarium\_fish" or ``What is this? This is beaver" (with class names inserted as-is from the CIFAR-100 dataset), resulted in an accuracy of 44.6\%.
We modified some of the class names, separating words and using simpler forms where appropriate (for example, ``aquarium\_fish" to ``aquarium fish" and ``aeroplane" to ``plane").
We also prepended the appropriate article (a/an) in object classification tasks. The result was prompts of the form ``What is this? This is an aquarium fish" and ``What is this? This is a beaver."
These changes alone increased accuracy of CLIP on CIFAR-100 up from 44.7\% to 69.1\%.

We also found that alternative phrasing of the question could impact performance.
For example, on the Abstract VQA dataset, prompting with the form ``Is the dog asleep?" yielded an accuracy of 8.3\% in one Flamingo model, while prompting with the form ``Using the image, the answer to Is the dog asleep? is most likely" yielded an accuracy of 29.3\% for the same model.

For this reason, we evaluated all models on a range of prompt formulations appropriate for each dataset.
A full description of the class name alterations and prompt variations can be found the in Appendix~\ref{app:prompts}.

\begin{figure}[tb]
    \centering
    \includegraphics[width=0.5\linewidth]{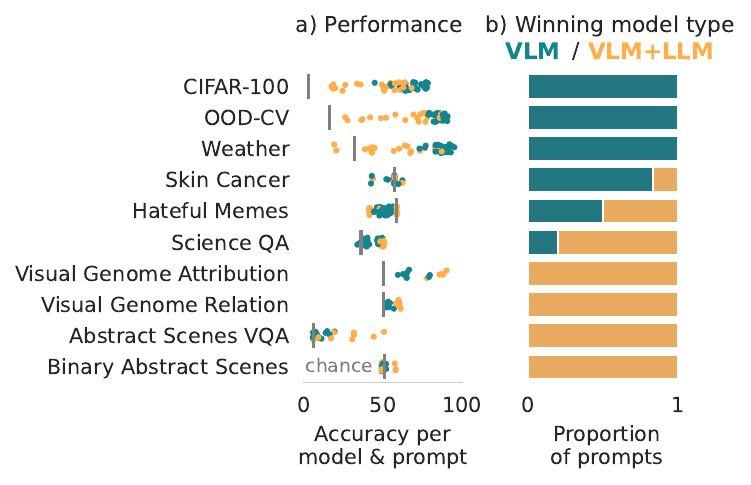}
    \caption{Input prompts affect model performance but mostly do not affect winning model type. a) Model performance for each model and prompt combination (markers) per dataset (y-axis). The accuracy varies across prompts, in particular for VLM+LLM models in recognition tasks (yellow markers in the first three rows). b) Proportion of times a VLM or VLM+LLM wins, across all prompts. Most datasets (y-axis) show a proportion of 1 for either VLM or VLM+LLM, indicating the winning model type is not affected by the specific prompt.}
    \label{fig:prompt_consistency}
\end{figure}

\subsection{Results}

We analyze results on 462 experiments, including ten datasets, seven models and between two to nine prompt strategies per dataset. Results are displayed in Table \ref{tab:classification_accuracy} and Figure \ref{fig:prompt_consistency}. Table \ref{tab:classification_accuracy} reports the average accuracy among prompts and also the best accuracy among prompts, for each model and dataset. We highlight in boldface the best model per dataset, and underline the best model among each group of VLM+LLM and corresponding VLM.  In order to ensure that the winning models displayed in Table \ref{tab:classification_accuracy} are largely prompt-independent, in Figure \ref{fig:prompt_consistency} we depict the dependency of our results on the prompting strategy. Concretely, in Figure \ref{fig:prompt_consistency}a, we display the distribution of accuracy across prompt variations, indicating with different colors whether the model was a VLM or a VLM+LLM. As expected, there is substantial variability in accuracy among models across different prompt variations, sometimes more than $50\%$. Despite this high variability, Figure \ref{fig:prompt_consistency}b shows the proportion of times a VLM or VLM+LLM obtains higher accuracy for each prompt variation. We observe consistent model type across datasets, which highlight that the trends are prompt-independent.

The key trends that we observe by analyzing the results in Table \ref{tab:classification_accuracy} and  Figure \ref{fig:prompt_consistency} are twofold: 

\begin{enumerate}

\item For most datasets that require visual reasoning and outside knowledge, VLM+LLMs achieves superior accuracy (see Figure \ref{fig:prompt_consistency}b and last six columns in Table \ref{tab:classification_accuracy}). VLM+LLMs showcase superior capabilities in extracting high-level semantic information and adapting to conceptual understanding, which may be an expected result. Note that the superior performance of VLM+LLMs cannot be attributed to the fact that the VLM+LLMs contain more parameters to process text. Despite the fact that LiT leverages a text encoder with many more parameters than CLIP's, CLIP achieves superior results to LiT on visual tasks that require reasoning. Thus, parameter count alone does not explain the superior performance of Flamingo over CLIP.

\item  For most datasets of object and scene recognition, VLMs consistently outperform their corresponding VLM+LLMs (see Figure \ref{fig:prompt_consistency}b and the first four result column in Table \ref{tab:classification_accuracy}), i.e.\ BLIP outperforms PNP-VQA and CLIP outperforms Flamingo. 
Note that this result cannot be attributed to the possibility that VLM+LLMs might struggle with closed-ended evaluation, as VLM+LLMs still excel in closed-ended visual tasks that require reasoning, as discussed before in point 1. We also observe that comparing the two versions of Flamingo, i.e.\ with the LLMs with 3B and 9B parameters, provides some improvements in some cases, but it is not sufficient to outperform VLMs. All this suggests that the reduction in accuracy may be attributed to the way the LLM is integrated with the vision encoder.

\end{enumerate}

It is notable that all models struggle in some of these benchmarks, with accuracies close to chance, specifically for the Hateful Memes and Skin Cancer datasets. This highlights the need for further improvement in zero-shot visual tasks, despite the significant strides in recent years.

Regarding the best-performing model among all, there is no clear winner. For recognition datasets, CLIP and LiT dominate. In datasets that require a higher degree of reasoning and outside knowledge, Flamingo and PNP-VQA excel depending on the dataset. Thus,  exploring the integration of these models into a unified system becomes a key research direction, allowing us to leverage the strengths of each model for optimal performance in all scenarios. In the next section, we pursue this idea by introducing a computationally efficient solution that combines the strengths of all the models into one system.

\section{A Fix: Routing the Task to the Best Model}
\label{sec:methods}

\begin{figure}[tb]
    \centering
    \includegraphics[width=0.5\linewidth]{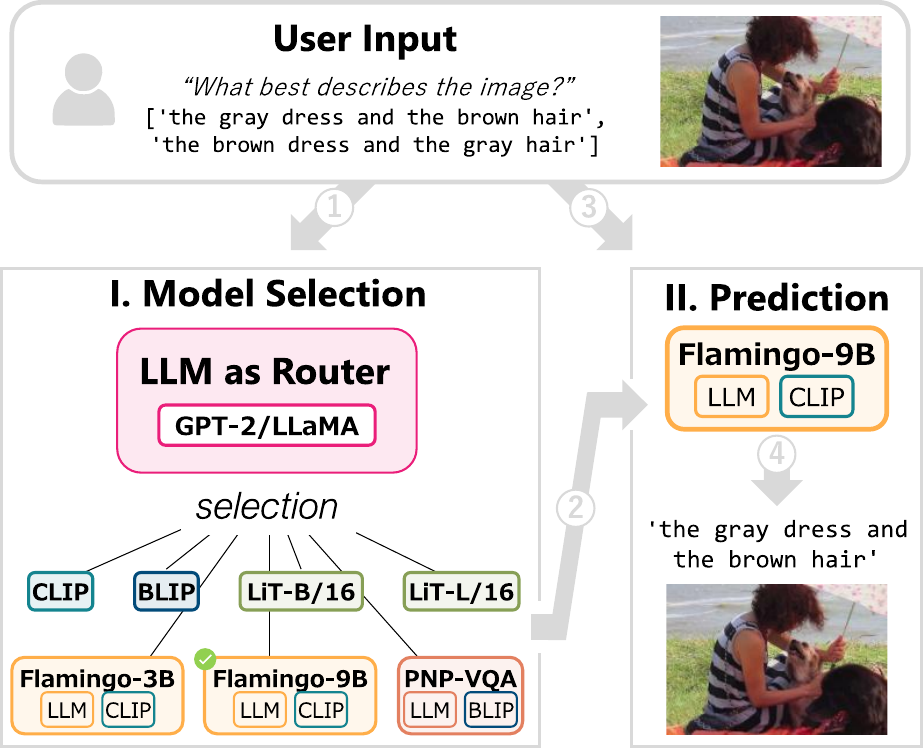}
    \caption{Our router approach where an LLM selects a suitable VLM/VLM+LLM to obtain high accuracy. The user input is first provided to the LLM router (1), which uses this information to select a model. The chosen model (2) is then provided with the user input (3) and generates a prediction (4).}
    \label{fig:our_planner_approach}
\end{figure}

Consider the following set up: a user provides a machine learning system with an image and a question about the image, and the response options. In Section~\ref{sec:observations} we have seen that VLMs and VLM+LLMs complement each other in the tasks they excel at. Since there is no single best model, we aim to design a system that can select the right model based on the user input. We propose to pass the user input to an LLM which then learns to route the input to the best model for the user's task. Figure~\ref{fig:our_planner_approach} shows an overview of this approach.

\subsection{Training an LLM to Act as a Router}
\label{sub:llmtraining}

Previous LLM-based model selection systems make use of human written descriptions of the model capabilities \cite{shen2023hugginggpt} or synthetically generated data \cite{patil2023gorilla}. In contrast, we use real model execution results calculated during our experiments (i.e.\ the experiments done to create Table~\ref{tab:classification_accuracy}). Importantly, this means our router learns from experience with actual data rather than some possibly inaccurate proxy as in previous works. 

After running experiments over all models, datasets and prompts we create a dataset of more than 2.5 million samples containing (i) an input prompt and image, (ii) a set of options to respond with, (iii) model performance. All of the input is represented in natural language. For the input image, we calculate simple image metadata in the form of image resolution and per channel mean and standard deviations which are prepended to the prompt.  The set of response options is a list of possible answers to the visual task. We find that the router performs best when the list of response options are not included as input to the LLM router, even though these are necessary to execute the predicted model for closed-ended evaluation. Per sample model performance is represented firstly by a boolean that is true if the sample is correctly classified by the model and false otherwise. Secondly, for each sample we additionally include average model performance over each dataset. Appendix~\ref{app:implementation} shows the full format of our data.

This initial dataset is then filtered by first removing any samples that use a prompt strategy that is not valid across all models (see Appendix \ref{app:prompts}). Following this, for a given input prompt and image we compare performance across all models to select the single best model, first by discarding any model that did not correctly classify the input and then selecting the model with the best average performance for the prompt on the dataset from the models remaining. In the case where all models incorrectly classify an image, we also select the model with best average performance. After this filtering process we are left with a training dataset containing approximately 280,000 training samples.

As we want our router to select a model based on the user's natural language query, we require an LLM. In order to keep the routing process relatively lightweight we pick a pre-trained GPT-2 \cite{radford2019gpt2} to use as a router. We then fine-tune GPT-2 taking the input prompt and image metadata as input, and the correct model as output, which is predicted as the next token after the input. The fine-tuning is done using the standard cross-entropy loss. Other implementation details are shown in Appendix \ref{app:implementation}. We also experiment with the router architecture by fine-tuning LLaMa \cite{llama} to assess whether our method is base model agnostic.

\begin{table}[t!]
\begin{scriptsize}
\centering
\begin{tabular}{@{}l@{\hspace{0.15cm}}c@{\hspace{0.15cm}}c@{\hspace{0.15cm}}c@{\hspace{0.15cm}}c|c@{\hspace{0.15cm}}c}
\toprule
& \multirow{2}{*}{\textbf{Chance}} & \multirow{2}{*}{\textbf{Average}} & \multirow{2}{*}{\textbf{Voting}} & \textbf{Router}  &  \multirow{2}{*}{\textbf{Oracle}} & \textbf{Upper} \\

{} & {} & {} & {} & {\textbf{(GPT-2)}}  & {} & \textbf{Bound} \\ 
\specialrule{\cmidrulewidth}{0pt}{0pt}
\textbf{CIFAR-100}         & 1.0\hspace{-1.4mm} & 60.9 & \textbf{{76.5}} & 60.7  & 72.7 & 92.3 \\
\textbf{OOD-CV}            & 16.1 & 81.4 & \textbf{{89.5}} & {{85.9}} &  89.3 & 97.5 \\
\textbf{Weather}          & 31.9 & 70.9 & \textbf{{84.1}} & \textbf{{84.1}} & 86.4 & 96.4 \\
\textbf{Skin Cancer}      & \textbf{{57.1}} & 41.7 & 40.8 & {{48.7}} &  50.0 & 86.3 \\
\textbf{Hateful Memes}    & \textbf{{58.7}} & 49.4 & 48.4 & 49.7 &  53.4 & 98.2 \\
\textbf{ScienceQA}       & 36.0 & 44.7 & 44.0 & \textbf{{57.7}} &  56.6 & 92.5 \\
\textbf{VG Attribution} & 50.1 & 75.3 & 80.6 & \textbf{{88.8}} &  91.6 & 99.8 \\
\textbf{VG Relation} & 50.2 & 55.8 & 57.9 & \textbf{{60.4}} &  60.4 & 97.3\\
\textbf{Abstract Scenes} & 5.7\hspace{-1.4mm} & 17.2 & 16.9 & \textbf{{42.3}}  & 42.3 & 70.2 \\
\textbf{B. Abstract Scenes} & 50.8 & 51.4 & 52.3 & \textbf{{58.6}} &  58.5 & 92.1\\
\specialrule{\cmidrulewidth}{0pt}{0pt}
\textbf{Average} & 35.7   & 54.9 & 59.1 & \textbf{{63.7}} &  66.1 & 92.9 \\
\textbf{Average (Weighted)} & 18.0   & 55.5 & 61.2 & \textbf{{64.9}} &  68.4 & 89.6 \\
\specialrule{\cmidrulewidth}{0pt}{1pt}
\specialrule{\cmidrulewidth}{0pt}{1pt}
\textbf{Cost (N. of Models)} & - &  $\mathbf{\mathcal{O}(1)}$  & $\mathcal{O}(N)$ & $\mathbf{\mathcal{O}(1)}$ &  $\mathcal{O}(N)$ & $\mathcal{O}(N)$ \\
\bottomrule
\end{tabular}
\caption{Performance comparison for different model selection baselines. Accuracy is averaged across prompt strategies. Each row evaluates a separately trained router which sees no samples from the listed evaluation dataset during training. As well as surpassing other methods in terms of accuracy our router also has constant, rather than linear, computational cost in the number of models. We report average performance over datasets as well as the average weighted by the different sizes of the datasets.}
\label{tab:performance_comparison}
\end{scriptsize}
\end{table}

\subsection{Evaluating on Held-Out Datasets}
\label{sec:results}

To test the generalization abilities of the router on new tasks, we test on unseen datasets in a cross-validation style. We train ten separate models, each time holding out one of the ten datasets which is then used to evaluate the performance of the router. Thus, by testing the LLM router on out-of-distribution datasets, we can properly evaluate its generalization capabilities.

We initially compare the performance of our LLM router against basic model selection techniques both in terms of accuracy on unseen data and computational cost, expressed as the number of models that must be run in order to use the technique. These model selection baselines are the following: 

\emph{Chance}--no models are run and a random label is assigned to each image. \emph{Average}--we report the average performance over all models, note that in the limit this is the same as randomly sampling a model each time a new input is received. \emph{Voting}--all models are run and the modal output is selected, this can be seen as a form of model ensemble. \emph{Oracle}--requires knowing the average accuracies for all models on the held-out dataset and then selecting the best performing model for a given input. \emph{Upper bound}--per image oracle that shows how often at least one of the available models correctly classifies a given input.

In Table~\ref{tab:performance_comparison} we see that our router outperforms all baselines, achieving the best performance on the largest number of datasets (seven of ten)  as well as the best average performance. Most notably our method outperforms the strongest baseline of ensemble voting, whilst being more computationally efficient at inference time. Ensemble voting requires executing every model on the inputs, making it linear in the number of models. On the other hand, our method only executes the single model chosen by the router, so is of constant order. Compared to the best baseline with the same computational cost (\emph{average}), our method achieves almost a ten-percentage point improvement on average. 
Regarding the baselines that require access to ground truth results on the held-out dataset (oracle and upper bound), our method is very close to the oracle performance, suggesting our router often picks the best model despite not having access to ground truth data. Finally, the upper bound column shows that by collating only a relatively small number of models we can theoretically achieve extremely high performance if the perfect model could be picked for every individual image, which indicates that there is still room for future improvements.

\subsection{Comparison with State-of-the-art}

We additionally compare against two modern state-of-the-art solutions that are applicable to our use case. Firstly HuggingGPT \cite{shen2023hugginggpt} which uses prompt design and in-context learning to get GPT-3 \cite{brown2020gpt3} to provide a model, or set of models, from the HuggingFace Library \cite{wolf2020transformers} to execute the task and then summarize the results. Secondly, we consider the newly released GPT-4V(ision)~\cite{openai2023gpt4, openai2023gpt4v-systemcard}. We use GPT-4V to evaluate a very recent approach which, motivated by scaling laws \cite{kaplan2020scaling, zhai2022scaling}, focuses on massive compute and data as a solution for improving visual understanding. Whilst technical details about GPT-4V are relatively limited we do know that this system is substantially more compute heavy than our relatively small LLM router and open-source models, any one of which can run on a single V100 GPU. Details of our implementations of these methods can be found in Appendix~\ref{app:baselines}.

Table~\ref{tab:baselines} compares our LLM routing approach against these baselines. We note that due to rate restrictions and high monetary costs (Appendix~\ref{app:baselines}) this table evaluates a total of 365 samples randomly selected (40 per dataset, except Weather and Skin Cancer, which have 24 and 21, respectively). Whilst this limited number of samples means we must be careful not to read too much into small differences in performance for individual datasets, some interesting trends emerge overall. 

When comparing to HuggingGPT, our approach outperforms HuggingGPT on nine of ten datasets and by more than ten percent absolute accuracy on average. Although HuggingGPT has access to many more models than our system, its reliance on text descriptions of the models and in-context learning means it fails to perform as well as our approach which is trained on actual model accuracy. What's more, due to the careful prompt engineering and in-context samples, queries to HuggingGPT run for several thousand tokens which at current pricing for GPT-3 (text-davinci-003) means we faced an inference cost of approximately \$0.05 per sample. 

When comparing to GPT-4V our method performs competitively with the average performance difference being less than one percent, this is despite our method requiring vastly less computational resources\footnote[3]{Very conservatively assuming GPT-4V is at least as big as GPT-3 this means it has a minimum of 175 billion parameters. In contrast our largest models use around 10 billion parameters.}. 

We also evaluated an LLM router with LLaMA as the LLM. However, we did not observe an improvement of accuracy, possibly because more training data is needed given LLaMA has more parameters than GPT-2. In Appendix~\ref{app:results} we show further results with the LLaMA router.

The success of our LLM router approach can be attributed to its specific training on a dataset that includes pairs of visual tasks and model accuracy. In contrast, other approaches rely on proxies like human-generated descriptions of the models.

\begin{table}[]
\begin{scriptsize}
\centering
\begin{tabular}{@{}l@{\hspace{-2mm}}cccc@{}}
\toprule
& \multirow{2}{*}{\textbf{HuggingGPT}} & \multirow{2}{*}{\textbf{GPT-4V}} & \textbf{Router} & \textbf{Router} \\
& & & \textbf{(GPT-2)} & \textbf{(LLaMA)} \\
\midrule
\textbf{CIFAR-100}             & 15.0        & 55.0   & \textbf{{60.0}} & \textbf{{60.0}} \\
\textbf{OOD-CV}                & 75.0        & \textbf{{82.5}}   & {{77.5}} & 72.5 \\
\textbf{Weather}              & 62.5        & {{91.7}}   & \textbf{{95.8}} & 75.0 \\ 
\textbf{ScienceQA}            & \textbf{{57.5}}        & \textbf{{65.0}}   & 50.0 & 47.5 \\
\textbf{Skin Cancer}          & 38.1        & 38.1   & \textbf{{61.9}} & \textbf{{61.9}} \\
\textbf{Hateful Memes}        & 42.5        & 45.0   & \textbf{{65.0}} & \textbf{{65.0}} \\
\textbf{Visual Genome Attribution}       & 87.5        & \textbf{{90.0}}   & {{87.5}} & 85.0 \\
\textbf{Visual Genome Relation}          & \textbf{{65.0}}        & {{62.5}}   & 57.5 & 57.5 \\
\textbf{Abstract Scenes VQA}    & {{37.5}}       & \textbf{{50.0}}   & 30.0 & 27.5 \\
\textbf{Binary Abstract Scenes} & 50.0        & \textbf{{77.5}}   & {{62.5}} & 57.5 \\
\midrule
\textbf{Average}              & 53.1        & \textbf{{65.7}}   & {{64.8}} & 60.9 \\
\textbf{Average (Weighted)}              & 53.4        & \textbf{{66.0}}   & {{63.6}} & 60.3 \\
\bottomrule
\end{tabular}
\caption{Comparison against state-of-the-art models on a separate smaller test set. Our method outperforms HuggingGPT and is competitive with GPT-4V with much lower computational cost.}
\label{tab:baselines}
\end{scriptsize}
\end{table}

\subsection{Analyses and Ablations}
In this section we further analyze our results, first by exploring how the router makes decisions on held-out datasets and then by conducting an ablation study where we experiment with the input information provided to the router. 

\begin{figure}[tb]
    \centering
    \includegraphics[trim={0.5cm 0 2cm 0},clip,width=0.8\linewidth]{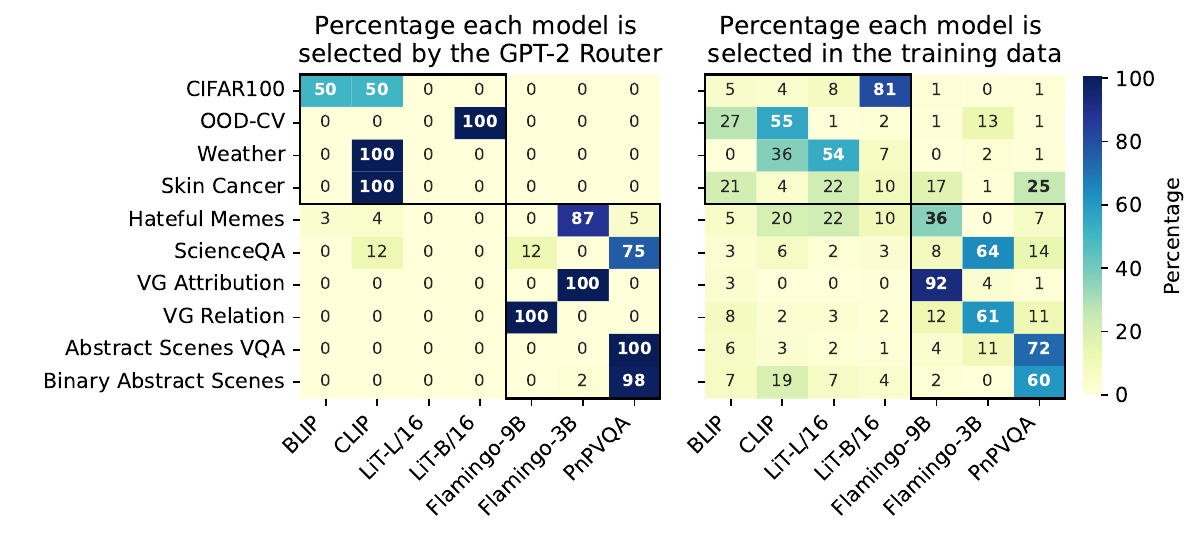}
    \vspace{-6mm}
    \caption{Heat maps of the LLM router's model selection distribution (left) and the distribution of models in the training data (right).}
    \label{fig:router_selection_vs_oracle}
\end{figure}

To explore how the router makes decisions we plot how often each model is selected by the router for each held-out dataset in Figure~\ref{fig:router_selection_vs_oracle} (left). We can then compare this to how often each model is used for each dataset in the training data in Figure~\ref{fig:router_selection_vs_oracle} (right). From this figure, we obtain the following insights about the decision-making process of the router. If we consider the dataset OOD-CV, we see that LiT-B/16 is always selected by the router. Note that this is the best model for CIFAR-100. Since OOD-CV is not seen during the training of the router, results suggest that the router perceives the unseen OOD-CV data is most similar to the input from CIFAR-100 and as a result, uses the best model for CIFAR-100 when presented with OOD-CV samples. We see similar patterns for other datasets like Visual Genome Attribution and Relation. This figure shows that the router can transfer knowledge between similar datasets due to the similarity in prompts for similar tasks. 


\begin{table}[t]
\begin{scriptsize}
\centering
\begin{tabular}{@{\hspace{-0.1mm}}l@{\hspace{-0.mm}}ccccc@{}}
\toprule
& \multicolumn{2}{c}{\textbf{w/ MD}} & \multicolumn{2}{c}{\textbf{w/o MD}} & \textbf{InD w/ MD} \\ \cmidrule(lr){2-3} \cmidrule(lr){4-5} \cmidrule(lr){6-6}
& w/ RO & w/o RO & w/ RO & w/o RO & w/o RO \\ \midrule
\textbf{CIFAR-100} & 56.8 & \textbf{60.7} & 56.6 & \textbf{60.7} & 72.7 \\
\textbf{OOD-CV} & \textbf{85.9} & \textbf{85.9} & \textbf{85.9} & \textbf{85.9} & 89.4 \\
\textbf{Weather} & 85.2 & 84.1 & \textbf{86.4} & 73.9 & 75.0 \\
\textbf{Skin Cancer} & \textbf{48.7} & \textbf{48.7} & \textbf{48.7} & 47.4 & 48.7 \\
\textbf{Hateful Memes} & 49.0 & 49.7 & 49.9 & \textbf{50.4} & 53.7 \\
\textbf{ScienceQA} & 52.4 & \textbf{57.7} & 52.9 & 57.3 & 56.0 \\
\textbf{Visual Genome Attr.} & 88.8 & 88.8 & \textbf{89.0} & 88.8 & 88.8 \\
\textbf{Visual Genome Rel.} & \textbf{60.4} & \textbf{60.4} & \textbf{60.4} & \textbf{60.4} & 58.5 \\
\textbf{Abstract Scenes VQA} & 16.1 & \textbf{42.3} & 25.8 & 38.5 & 42.7 \\
\textbf{Binary Abstract Scenes} & 58.5 & 58.6 & 58.5 & \textbf{58.8} & 58.5 \\
\midrule
\textbf{Average} & 60.2 & \textbf{63.7} & 61.4 & 62.2 & 64.4 \\ 
\textbf{Average (Weighted)} & 57.3 & \textbf{64.9} & 59.9 & 63.9 & 68.6 \\ 
\bottomrule
\end{tabular}
\caption{Ablation Study. We experiment with different input formats for the LLM router testing with and without image metadata (MD) and with and without the set of possible response options  (RO). The first four columns evaluate on held-out datasets, for comparison, we also show the in-distribution case (InD) where all datasets are seen during training. Accuracy is averaged across prompt strategies. }
\label{tab:ablation_study_heldout}
\end{scriptsize}
\end{table}

We additionally provide an ablation study using the GPT-2 router in Table \ref{tab:ablation_study_heldout}. We evaluated different input formats for the router (i) with or without the response options (RO) (ii) with or without image metadata (MD). Firstly, eliminating response options from the input prompt largely improves the performance. Thus, although we expected that including response options  would help to find similar datasets in the training data, it seems that response options  hinder rather than help. One possible explanation is that the response options are very specific to each dataset and act as shortcuts that lead to overfitting rather than generalization across datasets.

Recall that as input to the LLM router, we include simple metadata from the image (see section~\ref{sub:llmtraining}). Table \ref{tab:ablation_study_heldout} shows (a small) improvement due to the use of the image metadata. The dataset which has the largest performance gap when metadata is ablated is Weather (84.1\% vs.\ 73.9\%). This difference in performance suggests there may be image-level (rather than dataset-level) patterns that can be used to improve routing; an interesting avenue for further investigation is multimodal routers with stronger image processing abilities. See Appendix~\ref{app:results} for a more in-depth analysis.

Finally, in Table \ref{tab:ablation_study_heldout}, we show in-distribution performance, where the router is trained on all datasets and evaluated on held-out samples from datasets it has seen during training. To do so, we created an 80/10/10 train/validate/test split. Here, we see that even when datasets are held-out the router performs similarly to the in-distribution case, suggesting our router has learned to generalize well across datasets. A similar ablation study for LLaMa along with further heat maps are shown in Appendix \ref{app:results}.

\section{Related Work}
\label{sec:related_work}


\noindent\textbf{Foundation Models: VLMs \& LLMs.} Foundation models \cite{bommasani2021opportunities} have changed the landscape of vision and language, with large image \cite{kirillov2023segany, zou2023segment, rombach2021highresolution}, language \cite{brown2020gpt3, chowdery2023palm, touvron2023llama2} and explicitly multimodal \cite{minigpt4, li2023blip2, openai2023gpt4v-systemcard, chen2023pali, alayrac2022flamingo, driess2023palme} transformer based models \cite{vaswani2017attention} creating new capabilities in terms of generalization and low-shot performance \cite{bubeck2023sparks}. Of these various models, our work is most concerned with Vision Language Models \cite{radford2021clip, li2022blip, zhai2022lit} and methods that combine VLMs with LLMs \cite{alayrac2022flamingo, tiong2022pnp, huang2023kosmos}.

Some recent intriguing findings highlight the nuanced relationship between LLMs and VLMs \cite{roth2023waffling,alayrac2022flamingo,zhang23pretrained}. In the study by Roth et al.\cite{roth2023waffling}, CLIP evaluated on images with randomly generated query texts demonstrates comparable performance with CLIP evaluated on images with LLM-generated query texts. This is further evidence, complementary to ours, that how best to use LLMs to improve VLMs is remains an open question.
Meanwhile, a careful examination of the Flamingo paper, reveals that it lags behind state-of-the-art VLM models in ImageNet and  Kinetics700 classification tasks as shown in Appendix B in~\cite{alayrac2022flamingo}. 
This trend can also be seen by comparing Tables 5 and 19 in the results of PaLI \cite{chen2023pali}.
 Our findings makes this observation explicit and general. By focusing on the difference in performance of VLM+LLMs and their corresponding VLMs on classification tasks we reveal VLM+LLMs' limitations  across multiple datasets and models and then, we propose a solution to it.

\noindent\textbf{Routing, Model Selection \& Augmented LLMs.} Model selection is a standard problem in machine learning \cite{raschka2018model} but is most commonly examined in the case where validation data is used to select the best single model by estimating test error for future i.i.d. test data. In our case we instead use other datasets to learn a model which can dynamically select the best model for a given task, which we refer to as routing. Most similar to our work is concurrent work on routing for language benchmarks \cite{shnitzer2023routing}, which has interesting links to meta-learning \cite{vilalta2002perspective}. Yet, the results of Shnitzer et al.\cite{shnitzer2023routing} are limited to only natural language tasks.

Many tool augmented LLMs \cite{parisi2022talm, schick2023toolformer, mialon2023augmented, thoppilan2022lamda, viper}, that is LLMs that can reference external tools, can be viewed as performing routing to enable tasks beyond their initial training data. Several systems have been proposed to allow LLMs to reference external machine learning models however, these works are either aspirational in nature \cite{liang2023taskmatrix}, use synthetic data itself generated by an LLM \cite{patil2023gorilla}, or rely on prompt engineering and in-context learning \cite{shen2023hugginggpt, gupta2023visual}. In contrast, our method for routing learns cost-effective strategies from accuracies obtained by running and evaluating models, which serve as training data for the LLM router.

\section{Conclusions}
\label{sec:conclusion}

We found that VLM+LLMs were better at advanced reasoning and knowledge tasks but were worse on object and scene recognition tasks when compared with VLMs with the same vision encoder.
To address this issue, we proposed the use of a low-resource LLM that serves as a router, intelligently selecting the most suitable model among VLMs and VLM+LLMs for each visual task query. This LLM router strategy achieved higher accuracy than other model selection and ensemble baselines, and is superior or on par with several other state-of-the-art approaches, like HuggingGPT and GPT-4Vision, which use far more computational resources. 
Our LLM router's effectiveness is rooted in its specialized training on a corpus of examples, consisting of pairs of visual tasks and model accuracy. In contrast, competing approaches use proxies such as human-generated descriptions of the models.

The LLM router solution employs an LLM to address issues arising from the leveraging of LLMs to improve VLMs, i.e.\ an LLM is used to solve an issue that stems from another LLM. The effectiveness of the router solution indicates that LLMs alone have the potential to always augment the VLM capabilities across all visual tasks. Therefore, future investigations aimed at leveraging LLMs for VLMs should focus on identifying more effective strategies for integrating them, as suggested by the success of our LLM router solution.




\section*{List of Contributions}
A. Cooper, K. Kato, C. Shih, I. Mason and X. Boix conceptualized the research with contributions from K. Takemoto, H. Yeh and H. Yamane; A. Cooper, C. Shih and I. Mason wrote teh code and ran the experiments in Section 2; K. Kato and I. Mason wrote the code and run the experiments in Section 3; K. Takemoto, T. Sunagawa, H. Yeh and J. Yamanaka and H. Yamane contributed to the code to run the experiments; H. Yamane and K. Vinken analyzed and visualized the data, with contributions from I. Mason and A. Cooper; H. Yamane imported the datasets with contributions from K. Takemoto;  A. Cooper, K. Kato, C. Shih, I. Mason, K. Vinken and X. Boix wrote the paper with contributions from K. Takemoto and  H. Yamane, H. Yeh; X. Boix ideated and supervised the research with contributions of I. Mason.













\bibliography{icml}

\newpage
\appendix
\onecolumn

\section{Datasets}
\label{app:datasets}

\begin{figure*}[tb]
    \centering
    \includegraphics[width=0.95\linewidth]{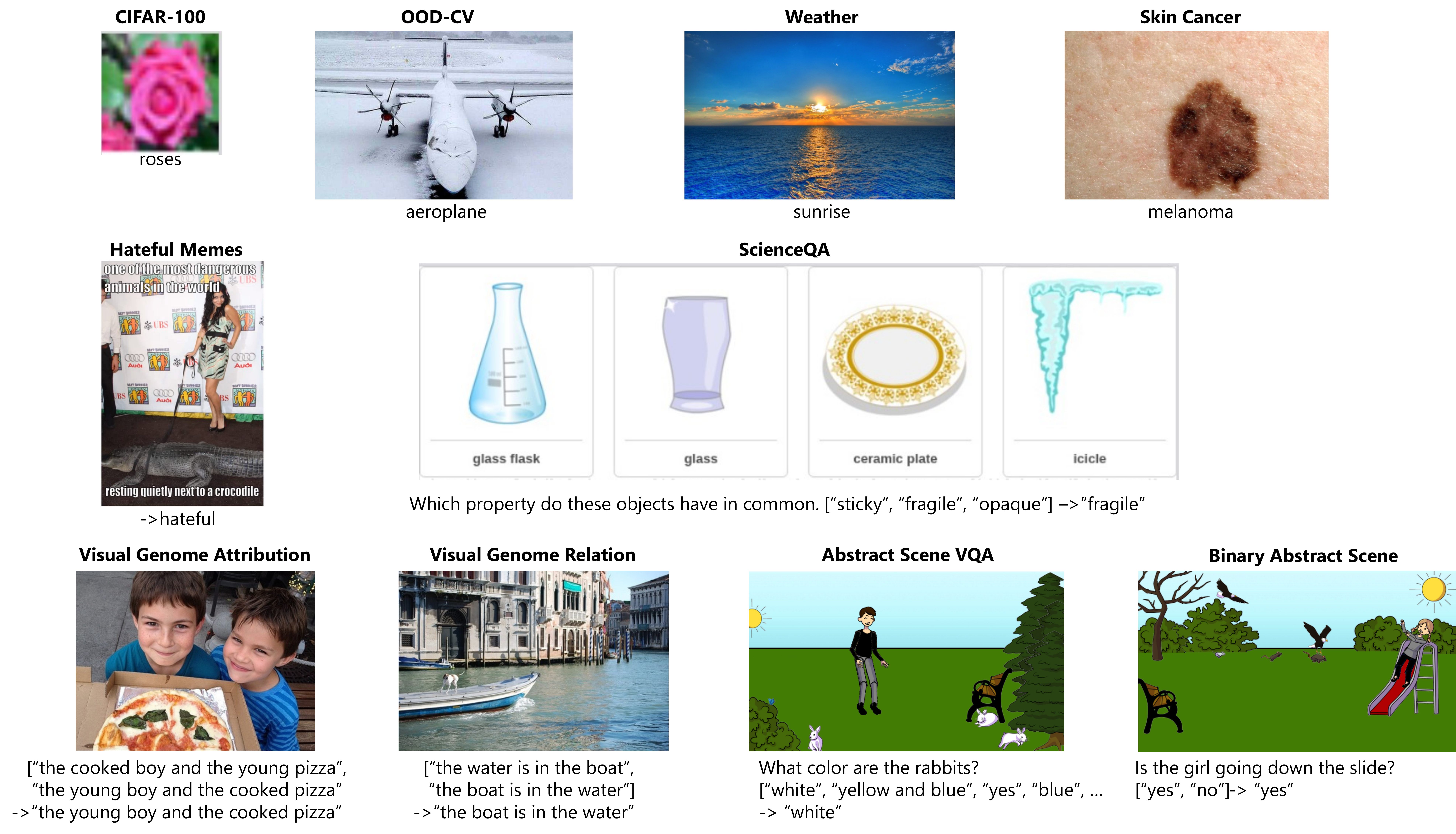}
    \caption{Example samples from the datasets used to evalute VLMs and VLM+LLMs.}
    \label{fig:dataset}
\end{figure*}

We utilize multiple datasets with distinct characteristics to evaluate the performance of our proposed model. The datasets can be divided into two categories, tasks of {\it Object \& Scene Recognition} (i.e., CIFAR-100, OOD-CV, Weather, and Skin Cancer) and those of {\it Visual Reasoning \& Outside Knowledge} (i.e., Hateful Memes, ScienceQA, Visual Genome Attribution, Visual Genome Relation, Abstract Scenes VQA, and Binary Abstract Scenes.). 
When evaluating zero-shot model performance we use the test sets of the datasets to reduce the chance that the models we evaluate have seen the images during training.
We show example images from datasets in Figure \ref{fig:dataset}.



\paragraph{CIFAR-100} The CIFAR-100 dataset~\cite{krizhevsky2009learning} consists of 60,000 32x32 pixel color images.
Each image contains an object from one of 100 classes.
This dataset is widely utilized for general object recognition tasks.
We adhere to the same train and test split as originally proposed.
The test split contains 10,000 images.

\paragraph{OOD-CV} OOD-CV~\cite{oodcv} is a benchmark dataset introduced to enhance the evaluation of vision algorithms' robustness in real-world scenarios.
The dataset includes out-of-distribution examples across 10 object categories, with variations in pose, shape, texture, context, and weather conditions.
It enables benchmarking models on tasks including image classification, object detection, and 3D pose estimation.
The dataset is composed of images from PASCAL3D+ and additional images collected and annotated by the creators.
We combine {\it phase-1} and {\it phase-2} for a total of 21,502 images in the test set.

\paragraph{Weather} The Multi-class Weather Dataset \cite{weather} is designed to test classification capabilities on images of various weather conditions.
It consists of a total of 1,125 images, categorized into four classes: Sunrise, Shine, Rain, and Cloudy.
The test split is 226 randomly sampled images (20\% of the dataset).

\paragraph{Skin Cancer} 
While prior studies have applied models like CLIP in the medical domain, specifically for diagnosing conditions such as COVID-19 and pneumonia from chest X-rays~\cite{medclip}, these images differ significantly from visible light RGB images.
The skin cancer dataset \cite{skin_cancer}, which focuses on melanoma detection natural images, is an excellent task candidate.
The dataset is maintained by the University of Waterloo.
It consists of images extracted from public databases DermIS and DermQuest, with response options healthy/cancerous and manual cropping to center the lesion in the image.
Since this is the smallest dataset, we sample 95\% of the dataset for evaluation, resulting in 196 images for testing.

\paragraph{Hateful Memes} 
The task in the Hateful Memes dataset~\cite{HatefulMemes} is to determine whether a meme is hateful or non-hateful. 
The dataset is designed to enforce multimodailty --- correctly classifying samples based on image or text alone is (essentially) impossible since individually the image or text may seem innocuous but combined they convey a mean or hateful message.
10,000 memes (image with text written on top) are collected/generated.
We combine {\it test\_seen} and {\it test\_unseen} as test dataset, resulting in 3,000 images in total.


\paragraph{ScienceQA} 
The ScienceQA dataset~\cite{ScienceQA} is a benchmark consisting of approximately 21,000 multimodal, multiple-choice questions spanning a diverse set of science topics, including abstract conceptual diagrams.
Each question is annotated with corresponding lectures and explanations, providing a rich context for assessment.
The dataset is designed to test the multi-hop reasoning ability of AI systems.
We exclude any samples from the test split that do not contain images, resulting in 2,017 images.

\paragraph{Visual Genome Attribution \& Relation}
The Visual Genome Attribution \& Relation dataset~\cite{VGAR} is crafted from the Visual Genome dataset~\cite{vg} and is a crucial part of the Attribution and Relation (AR) benchmark.
This benchmark is designed to assess the capabilities of vision and language models in understanding and processing complex relationships and attributes within images.
The dataset is structured to challenge models with tasks such as determining the correct order in relations (e.g., ``X relation Y'' vs ``Y relation X'') and accurately attributing properties to objects.
It includes 23,937 cases related to relations and 28,748 cases focused on attributes.
When tested on this dataset, existing models have shown deficiencies in the complex reasoning required to succeed at this task.
The test split contains 5,736 images for Visual Genome Attribution and 4,753 images for Visual Genome Relation. 

\paragraph{Abstract Scenes VQA and Binary Abstract Scenes}
This study utilizes Abstract Scenes, a part of the VQA v1 dataset~\cite{Antol_2015_ICCV}, which we refer to it as Abstract Scenes VQA.
The images are characterized by simplified, clip-art style representations, which tests the generalization capabilities of VQA models.
The tasks focuses on high-level semantics in vision and language, ensuring that both are essential for accurate responses.
We use 30,000 test images for Abstract Scenes VQA and 11,327 test images for Binary Abstract Scenes.

\section{Prompt Variations}
\label{app:prompts}

As discussed in Section \ref{prompting_choice}, varying the prompt had a significant impact on model performance.

Note that we employ closed-ended evaluation in this study, as discussed in Section \ref{closed-ended}.
Therefore, when we say we prompt the model, what we mean is that we provide it with a set of text prompts that consists of a single question combined with each of the options (from the set of response options).

\subsection{Prompt Grammar}
In order to methodically explore a range of different prompts, we outlined a small grammar to compose prompts.
The grammar is as follows:
\begin{itemize}
    \item \textit{key}: Datasets have several text fields, including the class name and additional text-context. One value that is used across datasets is \textit{class\_name}. When this value is used, $n$ prompts are generated, where $n$ is the number of classes in the dataset (for ex, 100 in the case of CIFAR-100) and each prompt is set to one of the class names. Some datasets have dataset-specific keys
    \item \textit{a\_a}n: An argument that adds an a/an appropriately to the key
    \item \textit{rename\_classes}: If the key text has a defined substitution, the key will be swapped for its rename
    \item \textit{arg}: Can be one of [\textit{a\_an}, \textit{rename\_classes}]
    \item \{[\textit{arg}]* $\vert+$ \textit{key}\}: The tag is fully described by zero or multiple \textit{arg}s, a $\vert$ character (which can be omitted if no args are specified) and the key
\end{itemize}
Additional text can surround the tags.

Further, each dataset defines a set of questions and options.
If a set of options is not defined, this means that the options are provided by the dataset for each sample.
A prompt is formed by concatenating a question to an option formulation.
Sometimes we leave the question empty to test whether the class name alone is the best prompt.
The cartesian product between all question and option formulations defines the full set of prompts.
Below we outline the prompts, options and additional text fields for each dataset.

\subsection{Prompts for each dataset}

\paragraph{CIFAR-100}
\begin{itemize}
    \item Question Formulations
    \begin{itemize}
        \item ``"
        \item ``This is "
        \item ``What is this? This is "
    \end{itemize}
    \item Option Formulations
    \begin{itemize}
        \item ``\{\textit{class\_name}\}"
        \item ``\{\textit{rename\_classes} $\vert$ \textit{class\_name}\}"
        \item ``\{\textit{a\_an}, \textit{rename\_classes} $\vert$ \textit{class\_name}\}"
    \end{itemize}
    \item Class Renames
    \begin{itemize}
        \item aquarium\_fish: aquarium fish
        \item pickup\_truck: pickup truck
        \item lawn\_mower: lawn mower
        \item sweet\_pepper: pepper
        \item maple\_tree: maple
        \item oak\_tree: oak
        \item palm\_tree: palm
        \item pine\_tree: pine
        \item willow\_tree: willow
    \end{itemize}
\end{itemize}

\paragraph{OOD-CV} Has the same question and option formulations as CIFAR-100
\begin{itemize}
    \item Class Renames
    \begin{itemize}
        \item aeroplane: plane
        \item diningtable: table
    \end{itemize}
\end{itemize}

\paragraph{Weather}
\begin{itemize}
    \item Question Formulations
    \begin{itemize}
        \item ``"
        \item ``It is "
        \item ``The weather is "
    \end{itemize}
    \item Option Formulations
    \begin{itemize}
        \item ``\{\textit{class\_name}\}"
        \item ``\{\textit{rename\_classes} $\vert$ \textit{class\_name}\}"
    \end{itemize}
    \item Class Renames
    \begin{itemize}
        \item Sunrise: sunrise
        \item Cloudy: cloudy
        \item Shine: sunny
        \item Rain: rainy
    \end{itemize}
\end{itemize}

\paragraph{Skin Cancer}
\begin{itemize}
    \item Question Formulations
    \begin{itemize}
        \item ``"
        \item ``This is "
        \item ``This skin is "
    \end{itemize}
    \item Option Formulations
    \begin{itemize}
        \item ``\{class\_name\}"
        \item ``\{\textit{rename\_classes} $\vert$ \textit{class\_name}\}"
    \end{itemize}
    \item Class Renames
    \begin{itemize}
        \item melanoma: cancerous
        \item notmelanoma: healthy
    \end{itemize}
\end{itemize}

\paragraph{Hateful Memes}
Below, the key \textit{text} is provided by the dataset for each sample.
It is the text-string of the text written on the meme.
\begin{itemize}
    \item Question Formulations
    \begin{itemize}
        \item ``"
        \item ``\{\textit{text}\}. "
        \item ``\{\textit{text}\}. This meme is "
        \item ``This is an image of a meme. It contains the text: \{\textit{text}\}. The meme is "
    \end{itemize}
    \item Option Formulations
    \begin{itemize}
        \item ``\{\textit{class\_name}\}"
        \item ``\{\textit{rename\_classes} $\vert$ \textit{class\_name}\}"
    \end{itemize}
    \item Class Renames
    \begin{itemize}
        \item not mean: nice
        \item mean: mean
    \end{itemize}
\end{itemize}

\paragraph{ScienceQA}
Below, the keys \textit{class\_question} and \textit{class\_hint} are provided by the dataset for each sample.
\textit{class\_question} is the question for each sample, while \textit{class\_hint} is some additional (unnecessary but helpful) context for each sample provided by the dataset.
\begin{itemize}
    \item Question Formulations
    \begin{itemize}
        \item ``"
        \item ``\{\textit{class\_question}\} "
        \item ``\{\textit{class\_hint}\} \{\textit{class\_question}\} "
        \item ``Question: \{\textit{class\_hint}\} \{\textit{class\_question}\} "
        \item ``\{\textit{class\_hint}\} Question: \{\textit{class\_question}\} "
    \end{itemize}
\end{itemize}

\paragraph{Visual Genome Attribution \& Relation}
Below, the key \textit{class\_question} is provided by the dataset for each sample.
It is the question for each sample.
\begin{itemize}
\item Question Formulations
    \begin{itemize}
        \item ``"
        \item ``\{\textit{class\_question}\} "
        \item ``This best describes the image: "
    \end{itemize}
\end{itemize}

\paragraph{Abstract Scenes VQA and Binary Abstract Scenes}
Below, the key \textit{class\_question} is provided by the dataset for each sample.
It is the question for each sample.
\begin{itemize}
\item Question Formulations
    \begin{itemize}
        \item ``"
        \item ``Question: \{\textit{class\_question}\} Answer: "
        \item ``Using the image, the answer to \{\textit{class\_question}\} is most likely "
    \end{itemize}
\end{itemize}

\subsection{Prompting PNP-VQA}
The implementation of PNP-VQA requires a question to serve as the prompt, separately from the options.
However for several datasets, we tried prompts that have an empty question, denoted with ``".
In these cases, we provide the empty string to PNP-VQA, to ensure that evaluation is the same across models.
However PNP-VQA cannot provide a response in these cases.
For the experiments in Table~\ref{tab:classification_accuracy} we use all the prompts described in this section.
When training the planner, to ensure all models have the same number of prompts and to avoid prompt/model dependence, we remove this question from further analysis for all models.

\section{Baseline Implementation}
\label{app:baselines}

This section describes how the baselines GPT-4V and HuggingGPT were implemented to create a fair comparison with our LLM router.

\subsection{GPT-4V(ision)}

GPT-4V(ision) was released for developer access on 11/6/23. Our experiments took place between 11/13/23 and 11/16/23. During this period GPT-4V was in preview, meaning organizations could only make 100 API requests per day, which limited the number of images we were able to test with it.

Further, though the API documentation provides for options that might be used to approximate close-ended evaluation, we were unable to make these parts of the API perform as documented for GPT4-V.
We therefore took an alternative approach to simulate closed-ended evaluation.
We provided the following system prompt:
\\

\noindent ``Complete the prompt from the list called OPTIONS, enclosed by [ and ]. Your response can only include a single element from this list"
\\

\noindent and in the user message, we provided the prompt and the closed set of response options within brackets. For each dataset we pick the prompt that is valid for all models and achieved the highest accuracy in the experiments to create Table~\ref{tab:classification_accuracy}.

To evaluate GPT-4V each completion was determined to be one of:
\begin{itemize}
    \item \textit{Correct Prediction}: if the response of the model contained only one element from OPTIONS and it was the correct label.
    \item \textit{Incorrect Prediction}: if the response of the model contained only one element from OPTIONS and it was an incorrect label.
    \item \textit{No Guess}: if the response contained no elements from OPTIONS
    \item \textit{Multiple Guesses}: if the response contained multiple elements from OPTIONS
\end{itemize}
To calculate the accuracies in Table~\ref{tab:baselines} we considered \textit{Incorrect Predictions}, \textit{No Guess} and \textit{Multiple Guesses} as errors.
This means the accuracy was calculated as the number of correct predictions divided by the total number of samples.

To more accurately measure the performance of GPT-4V we allow ten completions from the model.
We select the most common prediction (correct or incorrect) from GPT-4V as the final prediction, only selecting \textit{No Guess} or \textit{Multiple Guesses} if all of the ten completions fail to make a valid prediction. 
This approximates choosing the most probable option from the closed set.

\subsection{HuggingGPT}

To run HuggingGPT we use the code made publicly available at \url{https://github.com/microsoft/JARVIS}, the code was pulled on 10/19/23, HuggingGPT uses text-davinci-003 as the LLM. Due to making multiple requests with long context prompts, we found HuggingGPT cost approximately \$0.05 per-sample, or around \$20 to calculate results on the small subset of data in Table~\ref{tab:baselines} (excluding experimentation and development requests). We evaluated HuggingGPT in the same manner as GPT-4V, but calculated only one completion per sample due to these costs.

For the most part we used the default settings provided in the repository, making the following changes to the prompting to encourage the system to select from the set of valid response options.

\begin{itemize}
    \item \textbf{System Prompt}:
    \begin{itemize}
        \item \textbf{Original}: ``\#4 Response Generation Stage: With the task execution logs, the AI assistant needs to describe the process and inference results.''
        \item \textbf{Ours}: ``\#4 Response Generation Stage: With the task execution logs, the AI assistant needs to provide the best completion for the prompt provided by the user.''
    \end{itemize}
    \item \textbf{User Message}:
    \begin{itemize}
        \item \textbf{Original}: ``Yes. Please first think carefully and directly answer my request based on the inference results. Some of the inferences may not always turn out to be correct and require you to make careful consideration in making decisions. Then please detail your workflow including the used models and inference results for my request in your friendly tone. Please filter out information that is not relevant to my request. Tell me the complete path or urls of files in inference results. If there is nothing in the results, please tell me you can't make it.''
        \item \textbf{Ours}: ``Yes. Please complete the prompt from the list called OPTIONS, enclosed by [ and ]. Your response can only include a single element from this list.''
    \end{itemize}
\end{itemize}

\noindent The user message is then additionally given the instruction: 
\\

\noindent ``Using the image and the options provided please complete the following prompt. \{prompt\}... OPTIONS: \{response options\}'', where \{prompt\} and \{response options\} are replaced with the relevant prompt and response options for the given sample. As with GPT-4V, for each dataset we pick the prompt that is valid for all models and achieved the highest accuracy in the experiments to create Table~\ref{tab:classification_accuracy}.
\\

We also found HuggingGPT was rarely able to actually access and run models from the Hugging Face Hub. Due to this limitation and to improve reproducibility we ran HuggingGPT in local mode, downloading all relevant implemented models. The models available locally to HuggingGPT were downloaded on 11/11/23. The full list of models we made available to HuggingGPT is:
\begin{itemize}
    \item nlpconnect/vit-gpt2-image-captioning
    \item lllyasviel/ControlNet
    \item lllyasviel/sd-controlnet-canny
    \item lllyasviel/sd-controlnet-depth
    \item lllyasviel/sd-controlnet-hed
    \item lllyasviel/sd-controlnet-mlsd
    \item lllyasviel/sd-controlnet-openpose
    \item lllyasviel/sd-controlnet-scribble
    \item lllyasviel/sd-controlnet-seg
    \item runwayml/stable-diffusion-v1-5
    \item damo-vilab/text-to-video-ms-1.7b
    \item microsoft/speecht5\_asr
    \item JorisCos/DCCRNet\_Libri1Mix\_enhsingle\_16k
    \item espnet/kan-bayashi\_ljspeech\_vits
    \item facebook/detr-resnet-101
    \item microsoft/speecht5\_hifigan
    \item microsoft/speecht5\_vc
    \item openai/whisper-base
    \item Intel/dpt-large
    \item facebook/detr-resnet-50-panoptic
    \item facebook/detr-resnet-50
    \item google/owlvit-base-patch32
    \item impira/layoutlm-document-qa
    \item ydshieh/vit-gpt2-coco-en
    \item dandelin/vilt-b32-finetuned-vqa
    \item lambdalabs/sd-image-variations-diffusers
    \item facebook/maskformer-swin-base-coco
    \item Intel/dpt-hybrid-midas
    \item Salesforce/blip-image-captioning-large
    \item facebook/maskformer-swin-large-ade
    \item microsoft/beit-base-patch16-224-pt22k-ft22k
    \item openai/clip-vit-large-patch14
    \item deepset/roberta-base-squad2
    \item google/tapas-base-finetuned-wtq
    \item distilbert-base-uncased-finetuned-sst-2-english
    \item gpt2
    \item Dataset: Matthijs/cmu-arctic-xvectors
\end{itemize}

\section{Implementation detail}
\label{app:implementation}

In this section, we provide additional implementation details for evaluations and experiments.

\subsection{Closed-Ended Evaluation}
In Section \ref{closed-ended}, we motivated our use of closed-ended evaluation.
As mentioned, the implementation of closed-ended evaluation differed across models.
Here we outline the implementation in various model architectures.
This evaluation is computed for each sample in the dataset, where each sample is an image and the corresponding set of text prompts ($o \in O$): question and response options.

\paragraph{Constrastive Models} This group includes CLIP, BLIP, and LiT.
The model has two different encoders, one for images and one for text.
For each sample, we extract the $N$-dimensional image embedding, $v_i \in \mathbb{R}^{N}$, and the text embedding matrix, $V_O \in \mathbb{R}^{|O| \times N}$, where rows of $V_O$ are the $N$-dimensional embedding of the prompts $o \in O$.
We then take the prediction to be the option indexed by $\text{argmax}(v_i^T \cdot V_O)$.

\paragraph{Language Models} This group includes PNP-VQA, and Flamingo.
Models of this type can return a log-probability for the next token, across all tokens in the vocabulary, given the preceding tokens.
These probabilities are generally used for generating new text, but they can also be used for closed-ended evaluation.
For each prompt, $o \in O$, we sum the log probabilities for each of the tokens in the prompt. 
The prediction is then the response option used to create the prompt which has the highest probability.

\subsection{Data Format}
Section~\ref{sub:llmtraining} describes the creation of training data for the LLM router. After this process a single sample in the dataset looks like

\begin{verbatim}
[img]dim::(270,317,3)ave::(23.1,31.8,
46.2)std::(15.1,11.5,10.9)[prompt]What
is this? This is ;;;[`a car', `a sofa',
`a train', `a table', `a chair', 
`a boat', `a plane', `a motorbike',
`a bus', `a bicycle'[SEP]model::
clip[response]correct::True;;;
avg_accuracy::0.88238 
\end{verbatim}

The values following the [img] marker are the resolution of the image along with per-channel mean and standard deviation of the pixel values. Then the prompt that should be completed is provided along with the set of response options (a car, a sofa etc.). These components are the possible inputs to the router that we experiment with. After the [SEP] token is a label for the model (CLIP). The final part is the boolean that marks whether or not the model gave the correct response for the image in questions as well as the average accuracy for the model on the dataset the image is from. The boolean and the average accuracy are used to select the best model for each image which is then used to train the LLM router as described in Section~\ref{sub:llmtraining}.

\subsection{GPT-2 Implementation}
For GPT-2, we used the pre-trained model from
\url{https://huggingface.co/gpt2}
The optimizer was Adam with an initial learning rate of $2 \times 10^{-4}$. The batch size was 1, we trained for 5000 iterations, after every 1000 iterations, an evaluation was done and the checkpoint with the best result was used as the final model. 

\subsection{LLaMA Implementation}
For LLaMA, we used a pre-trained model from \url{https://huggingface.co/baffo32/decapoda-research-llama-7B-hf}.
LoRA with rank 8 and model precision of fp16 were applied to save computational cost. We optimized models with the AdamW optimizer with an initial learning rate of $1 \times 10^{-3}$. The batch size was 2. we trained for 500 iterations, an evaluation was done every 100 iterations and the checkpoint with the best result was used as the final model.

\section{Further Results}
\label{app:results}

\subsection{Further Ablation Studies for GPT-2}
First, we examine the results of training the planner on all datasets and evaluating on unseen samples from those datasets, see Table~\ref{tab:ablation_study_ind}. This is slightly different from holding out individual datasets as was done in the main text, instead this is `in-distribution' evaluation. When compared with Table~\ref{tab:ablation_study_heldout} we see that the router is only slightly better in this case despite being evaluated in-distribution. This shows that the LLM router described in the main text is able to generalize well to held-out datasets.

\begin{figure}[tb]
    \centering    \includegraphics[width=0.8\linewidth]{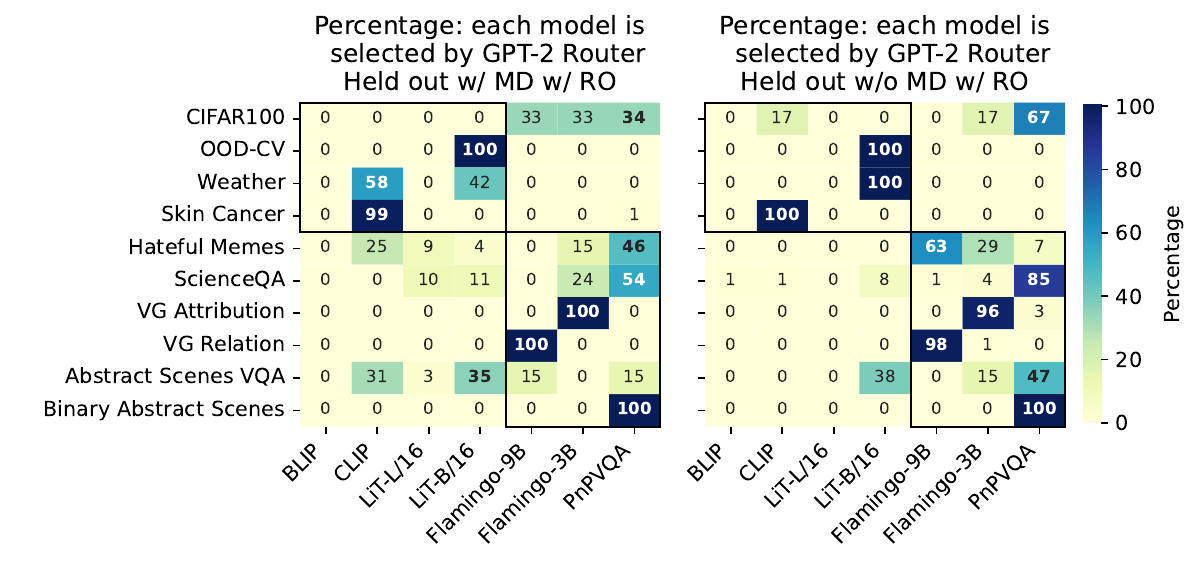}
    \caption{GPT-2 Router,  model selection across datasets and prompt strategies. Left: with metadata and with response options vs Right: without metadata and with response options. }    \label{fig:router_selection_vs_w/o_meta_w_label}
\end{figure}

\begin{figure}[tb]
    \centering    \includegraphics[width=0.8\linewidth]{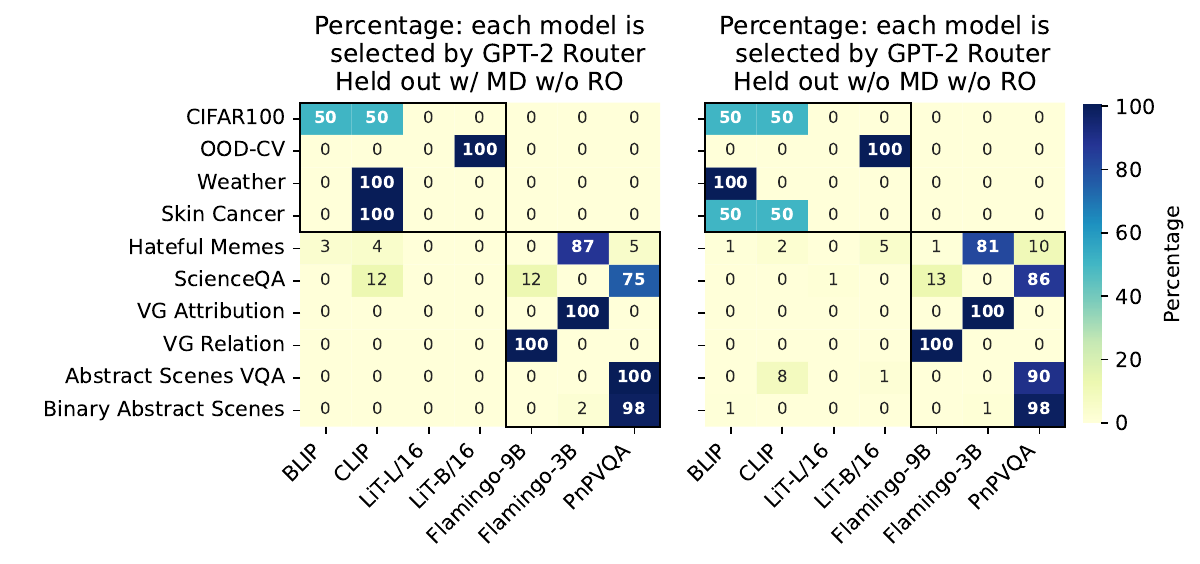}
    \caption{GPT-2 Router,  model selection across datasets and prompt strategies. Left: with metadata and without response options vs Right: without metadata and without response options.}    \label{fig:router_selection_vs_w/o_meta_wo_label}
\end{figure}

We also include heatmaps for the GPT-2 Router performance on held-out datasets with and without metadata and with and without response options, see Figures \ref{fig:router_selection_vs_w/o_meta_w_label}, \ref{fig:router_selection_vs_w/o_meta_wo_label}.
Conclusions are the same as in the ablation study in Table~\ref{tab:ablation_study_heldout}. Note that routers trained with response options are more likely to choose an inferior model type (VLM or VLM+LLM) for a given dataset type (classification or reasoning). Instead, relying only on the prompts seems to make it easier for the router to determine the task type. 

Next, we compare the routers trained with and without metadata, and without the response options. 
The router always selects CLIP as the best model when trained with metadata (Figure~\ref{fig:router_selection_vs_w/o_meta_wo_label}, left), but when trained without metadata the router chooses CLIP and BLIP with equal probability (Figure~\ref{fig:router_selection_vs_w/o_meta_wo_label}, right).
Since CLIP is significantly better than BLIP on the Weather dataset, the router trained with metadata achieves higher performance.
In the training data (Figure~\ref{fig:router_selection_vs_oracle}, right), BLIP is selected by the router relatively often for the OOD-CV and Skin Cancer datasets. 
Since the images from Weather and Skin Cancer datasets look very different, it is likely they can be distinguished even with the simple metadata provided to the router, and therefore the router may learn to avoid using BLIP for such visually different images. 

\begin{table}[h]
\begin{scriptsize}
\centering
\begin{tabular}{@{}lllll@{}}
\toprule
& \multicolumn{2}{c}{\textbf{w/ MD}} & \multicolumn{2}{c}{\textbf{w/o MD}} \\ \cmidrule(lr){2-3} \cmidrule(lr){4-5}
& w/ RO & w/o RO & w/ RO & w/o RO \\ \midrule
\textbf{CIFAR100} & 72.7 & 72.7 & 71.5 & 71.5 \\
\textbf{OODCV} & 89.1 & 89.4 & 92.0 & 92.0 \\
\textbf{Weather} & 77.3 & 75.0 & 84.6 & 84.6 \\
\textbf{Skin Cancer} & 48.7 & 48.7 & 36.8 & 42.1 \\
\textbf{Hateful Memes} & 55.0 & 53.7 & 60.0 & 60.0 \\
\textbf{ScienceQA} & 58.6 & 56.0 & 56.0 & 59.5 \\
\textbf{VisualGenomeAttribution} & 91.6 & 88.8 & 94.5 & 94.5 \\
\textbf{VisualGenomeRelation} & 58.5 & 58.5 & 61.0 & 60.5 \\
\textbf{AbstractScenesVQA} & 42.3 & 42.7 & 46.5 & 45.0 \\
\textbf{BinaryAbstractScenes} & 58.5 & 58.5 & 54.0 & 54.5 \\
\midrule
\textbf{Average} & 65.2 & 64.4 & 65.7 & 66.4 \\
\textbf{Average (Weighted)} & 68.4 & 68.6 & 66.8 & 67.2 \\ \bottomrule
\end{tabular}
\caption{GPT-2: Ablation Study for the \emph{in-distribution} case where
all datasets are seen during training. We experiment with different input formats for the LLM router testing with and without image metadata (MD) and with and without the set of possible response options (RO). Accuracy is averaged across prompt strategies.}
\label{tab:ablation_study_ind}
\end{scriptsize}
\end{table}

\subsection{LLaMa Ablation}

\begin{table}[t]
\begin{scriptsize}
\centering
\begin{tabular}{@{\hspace{-0.1mm}}l@{\hspace{-0.mm}}cccc@{}}
\toprule
& \multicolumn{2}{c}{\textbf{w/ MD}} & \multicolumn{2}{c}{\textbf{w/o MD}} \\ \cmidrule(lr){2-3} \cmidrule(lr){4-5}
& w/ RO & w/o RO & w/ RO & w/o RO \\ \midrule
\textbf{CIFAR-100} & 52.5 & 64.5 & 58.8 & 66.1 \\
\textbf{OOD-CV} & 89.4 & 84.7 & 82.4 & 88.0 \\
\textbf{Weather} & 62.5 & 76.1 & 73.9 & 86.4 \\
\textbf{Skin Cancer} & 50.0 & 43.4 & 52.6 & 42.1 \\
\textbf{Hateful Memes} & 50.8 & 51.5 & 48.8 & 49.5 \\
\textbf{ScienceQA} & 58.5 & 52.2 & 54.8 & 54.8 \\
\textbf{Visual Genome Attribution} & 91.0 & 90.0 & 91.4 & 89.7 \\
\textbf{Visual Genome Relation} & 62.1 & 63.1 & 61.5 & 61.5 \\
\textbf{Abstract Scenes VQA} & 34.2 & 44.2 & 16.9 & 38.9 \\
\textbf{Binary Abstract Scenes} & 54.8 & 60.5 & 62.5 & 61.1 \\
\midrule
\textbf{Average} & 60.6 & 63.0 & 60.4 & 63.8 \\
\textbf{Average (Weighted)} & 62.4 & 65.8 & 57.1 & 65.9 \\
\bottomrule
\end{tabular}
\caption{Ablation Study for LLaMA (Held out). We experiment with different input formats for the LLM router testing with and without image metadata (MD) and with and without the set of possible response options (RO). Like GPT-2 it's evaluated on held-out datasets. Accuracy is averaged across prompt strategies.}
\label{tab:ablation_study_llama}
\end{scriptsize}
\end{table}

\begin{figure}[tb]
    \centering    \includegraphics[width=0.8\linewidth]{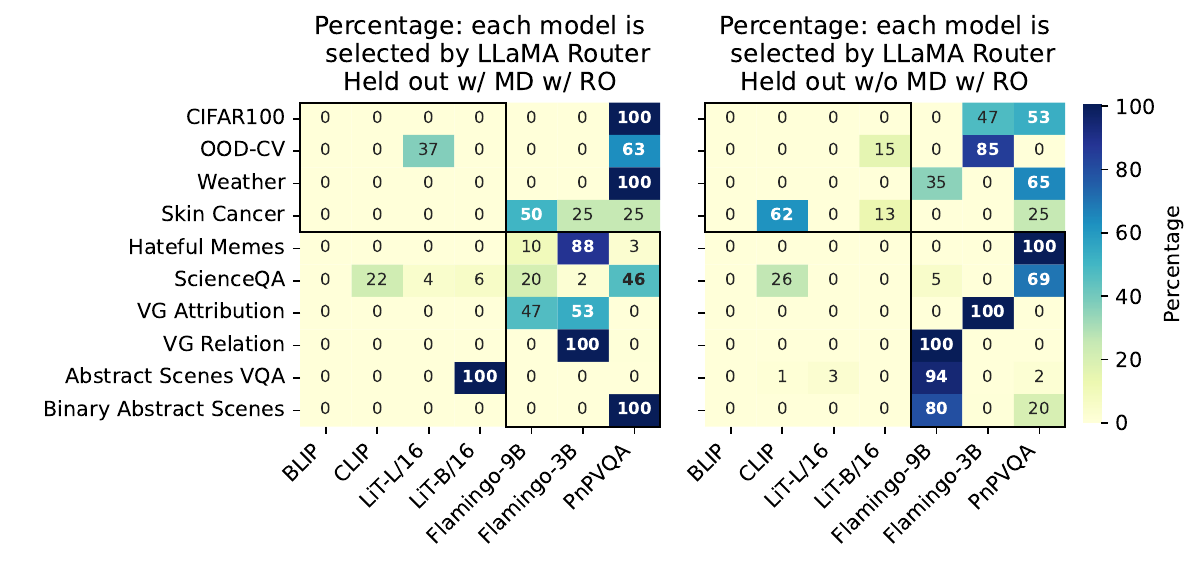}
    \caption{LLaMA Router, model selection across datasets and prompt strategies. Left: with metadata and with response options vs right: without metadata and with response options.}    \label{fig:llama_router_selection_vs_w/o_meta_w_label}
\end{figure}

\begin{figure}[tb]
    \centering    \includegraphics[width=0.8\linewidth]{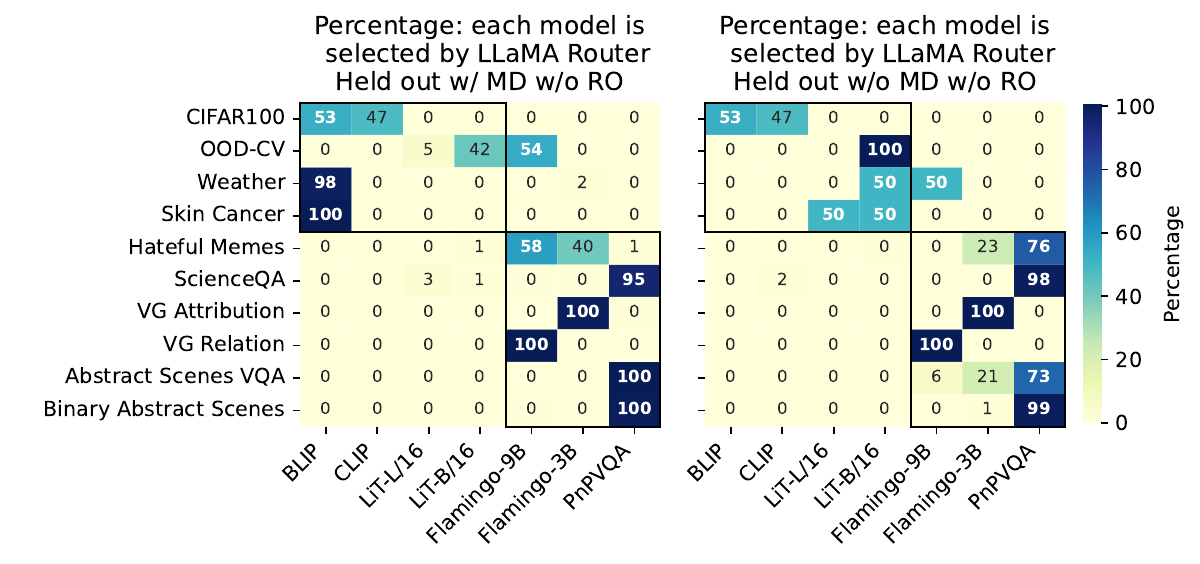}
    \caption{LLaMA Router, model selection across datasets and prompt strategies. Left: with metadata and without response options vs right: without metadata and without response options.}    \label{fig:llama_router_selection_vs_w/o_meta_wo_label}
\end{figure}

We also include an ablation and heatmaps for the LLaMA router performance on held-out datasets with and without metadata and with and without response options, see Table~\ref{tab:ablation_study_llama} and Figures \ref{fig:llama_router_selection_vs_w/o_meta_w_label} \& \ref{fig:llama_router_selection_vs_w/o_meta_wo_label}.
Similarly to the GPT-2 router, the LLaMA router trained without response options performed better. 
On the other hand, when comparing performance with and without metadata, training without metadata achieves marginally higher average accuracy.
As with GPT-2, by examining the heatmaps (Figures \ref{fig:llama_router_selection_vs_w/o_meta_w_label} \& \ref{fig:llama_router_selection_vs_w/o_meta_wo_label}), we see that including response options tends to lead to less optimal model selection and hence lower performance.

\end{document}